# BESLUTSSTÖDSSYSTEMET DEZZY - EN ÖVERSIKT

*Ulla Bergsten*
*Johan Schubert*
*Per Svensson*

**Försvarets Forskningsanstalt**
**Huvudavdelning för Vapensystem, Verkan och Skydd**
**102 54 Stockholm**

Inom ramen för FOA:s treåriga huvudprojekt Ubåtsskydd har delprojekt Informationssystem utvecklat demonstrationsprototypen Dezzy till ett beslutsstödssystem för hantering och analys av underrättelser om främmande undervattensverksamhet.

## MÅLSÄTTNING

Dezzy är i första hand ett system för situationsanalys, dvs ett system som med utgångspunkt från lagrade data om egen militärgeografi och egna ubåtsskyddsresurser kan ge stöd för bedömning av det aktuella hotet genom att underlätta kvantitativ analys av inkommande underrättelser.

I hotbedömningen ingår att så långt möjligt besvara frågor om antal hotenheter, enheternas typ, lägen, hastigheter och operativa uppgifter. Dessa bedömningar måste givetvis ske på en grund av mer eller mindre osäkra data. Därför är förmåga att hantera, presentera och dra slutsatser av osäkra data av avgörande betydelse.

En viktig del av denna förmåga är att kunna beräkna undre och övre gränser för en utsagas trovärdighet, givet att de fakta som är underlag för slutsatsen har känd tillförlitlighet.

Likaså är förmåga att göra känslighetsanalyser av stor betydelse, dvs att kunna beräkna i hur hög grad förändringar i en utvald delmängd av indata till en slutsats påverkar slutsatsens trovärdighet.

En fullständig situationsanalys torde vara mycket svår att åstadkomma. Huvuduppgiften för Dezzy har därför begränsats till att lösa följande delproblem:

(*i*) ***Antalet hotenheter***
Beräkna trolighetsgrad för de olika möjliga antalen.

(*ii*) ***Dessa enheters lägen***
Beräkna de områden där enheterna kan befinna sig, samt indela dessa områden i delområden med avseende på trolighetsgrad (*evidenskartor*).

(*iii*) ***Dessa enheters färdvägar***
Beräkna trolighetsgrad för de olika enheternas möjliga färdvägar fram till nuläget.

Eftersom situationsanalysen i denna tillämpning i hög grad måste baseras på analys av historiska underrättelsedata, snarare än på kunskap om militära doktriner eller militärtekniska fakta, är systemet försett med flexibla delsystem för lagring, analys och presentation av sådana data.

Dezzy's kärna är ett nyutvecklat system för formulering, värdering och presentation av hypoteser, grundade på osäkra observationsdata. Principerna för hantering av och slutsatsdragning ur osäkra data följer evidensteorin (*Dempster-Shafer*). Vi har valt denna metod därför att den enligt vår uppfattning lämpar sig bättre för denna form av osäkra data än alternativa metoder.

## EXEMPEL PÅ PROBLEM SOM KAN BEHANDLAS MED DEZZY

Vi tänker oss följande scenario: rapporter om ubåtsincident, sensorsignaler eller mänskliga observationer, flyter in till en rapportcentral. Varje rapport innehåller uppgifter om tidpunkt, läge, i vissa fall ubåtstyp, hastighet och/eller färdriktning etc. Dessutom är varje rapport försedd med klassificering av vilken tilltro man satt till rapporten.

### Sammanjämkning av rapporter

Systemet försöker urskilja ett mönster i en given rapportmängd. Systemet kan t ex konstruera hypoteser om antal ubåtar samt om vilka rapporter som härrör från respektive ubåt. Man kan här också ställa upp hypoteser om vilka rapporter som kan betraktas som falska. Ovanstående baserar sig på rapporternas trovärdighet, deras tidsmässiga och geografiska avstånd, förekomst eller avsaknad av sensorsignaler, framkomlighet (öar, olika djup) etc.

Hypoteserna är avsedda att användas tillsammans med operatörens egen kunskap om taktiska regler o d.

### Uppställande av hypoteser rörande lokalisering och uppgift för misstänkt ubåt

En misstänkt ubåt kan lokaliseras på flera sätt. Man kan beräkna troligaste uppehållsområden för en eller flera ubåtar baserade på en given rapportmängd, djupdata, sensoregenskaper mm. Man kan vidare beräkna troligaste färdväg genom rapportmängden för en eller flera ubåtar, samt göra en enkel prediktion av fortsatt färdväg.

### Interaktiv prövning av hypoteser uppställda av användaren mot en mängd rapporter från en incident

Antag att man förfogar över en mängd rapporter från samma incident. Mot denna vill man kontrollera ett antal hypotesers samklang med den givna rapportmängden.

Ändra antaganden, t ex om ubåtstyp eller genom att betrakta vissa rapporter som falska. Hur påverkar detta den förklaring som systemet ger?

### Efteranalys av incidenter

Efter det att ubåten har avlägsnat sig eller anträffats kartläggs inom vilka områden den kan ha (eller troligen har) befunnit sig.

Man kan plotta alla incidenter genom åren, eller en utvald mängd sådana, oberoende av tidpunkt men med exakt läge, för att få en bild av eventuella stråk och om vilka delar av svenskt vatten som utforskats. Man kan också göra statistiska sammanställningar av observationers fördelning samt presentera dem i en kartbild eller som diagram av olika slag.

Resultatet kan lagras i en incidentdatabas som sedan skall kunna användas för att analysera främmande ubåtars beteende och dra slutsatser om deras egenskaper, syften och taktik samt om vilken kännedom de kunnat skaffa sig om svenska vatten.

## ÖVERSIKT ÖVER DEMONSTRATIONSSYSTEMET

Demonstrationssystemet körs på tre samarbetande datorer och består av två huvudfunktioner, dels ett system (UBSSIM) som genererar rapporter från ett fiktivt scenario, dels av analyssystemet Dezzy.

### Datorutrustning

1. En Texas Instruments Explorer
   På denna maskin körs simuleringsprogrammet UBSSIM [Bråmå & Fransson, 1989] som genererar rapporter från en fiktiv incident samt levererar dem till:

2. En Vax-dator med ansluten grafisk presentationsutrustning av fabrikat RasterTech
   Här körs systemet Gral [Hedin, 1989], vilket möjliggör kartpresentation med hög kvalitet och funktionalitet. Vax-systemet skickar också rapporterna vidare till:

3. En Apollo arbetsstation av typ DN3000
   På denna körs databashanteraren Cantor, presentationsprogrammet Groda, analysmodulen Dezzy-AM, menyhanteraren Waiter samt det kommunikationsprogram som knyter samman dessa olika system med varandra och med Gral.

### Systemet Dezzy sett från användaren

Användarinteraktionen är baserad på dynamiska menyer, multipla grafikfönster samt rena textfönster. Inmatning till systemet sker genom val i meny, via tangentbordet eller genom pekning i en kartbild.

Samtliga scenariedata i systemet lagras i en databas. Sådana data är, förutom incidentrapporter, främst lägen och räckvidder för egna sensorer.

Användaren av systemet har tillgång till databasen via en i systemet integrerad relationsdatabashanterare, Cantor [FOA, 1989a]. Med hjälp av Cantor kan man ställa även mycket komplicerade frågor mot databasen. Cantor kan också användas för bearbetningar, exempelvis för att variera data i de inmatade rapporterna. Sådana bearbetade data kan t ex användas för att studera parameterändringars effekt på analysresultaten.

Kärnan i systemet utgörs av en analysmodul, med vars hjälp vissa beräkningar som baserar sig på evidenskalkyl och geometri kan utföras. Dessa metoder beskrivs mer ingående i senare avsnitt.

Analysmodulen kallar vi Dezzy-AM.

Användaren har också tillgång till kvalificerade presentationsmetoder som bl a gör det möjligt att skapa kartbilder med overlay-möjligheter (lägespresentation) samt att producera diagram av olika slag. För lägespresentationen används Gral-systemet, medan grafisk presentation av statistiska sammanställningar kan göras m.h.a. det för Cantorsystemet utvecklade presentationssystemet Groda [FOA, 1989b].

Cantor, Gral och Groda är generella interaktiva system, som inom ramen för Dezzy har integrerats och anpassats för en specifik tillämpning. De olika systemen har integrerats m h a ett kommunikationsprogram som arbetar enligt principen meddelandesändning mellan processer. Användaren adresserar de olika programmen via en central menyhanterare, och behöver därför inte känna till de olika programsystemens användarspråk i detalj.

Eftersom menyerna kan förändras av användaren (genom redigering av en menyfil) så kan nya funktioner mycket snabbt läggas in i Dezzy. För att kunna bygga ut systemet på detta sätt behöver man naturligtvis kunskap om såväl menyhanterarens som det via denna styrda systemets kommandospråk.

Samtliga nämnda programsystem har utvecklats av FOA.

**Datapresentationsfunktioner**

Ett viktigt syfte med Dezzy är att användaren skall kunna få data i databasen tydligt åskådliggjorda. För detta ändamål finns funktioner som kan utföra följande:

1. Plotta alla eller ett önskat urval av rapporter, t ex de som avser de senaste n timmarna, alla inom ett givet område, alla med en trovärdighet större än p etc (Cantor-Groda).

2. Genom att peka på en godtycklig rapport kan man få all information om denna utskriven på skärmen (Gral).

3. Skapa en karta som visar de fasta sensorernas placering inom angivet område. Vidare utritas räckviddsområden runt varje sensor som visar inom vilket område en ubåt troligen kan upptäckas (Gral).

4. En karta över möjliga stråk kan presenteras för att användaren skall kunna se hur rapporterna ansluter sig till dessa (Gral).

5. Ur försvarssynpunkt viktiga objekt kan efterfrågas och läggas in i lägeskartan (Gral).

6. Egna styrkors samt civila fartygs placering, kurs och fart kan läggas in i kartan (Gral).

**Analysfunktioner**

Ett antal analysfunktioner kan direkt utföras av Dezzys analysmodul. Dessa beskrivs nedan. Som redan nämnts kan ytterligare analysmetoder lätt läggas in i systemet. Vilka som helst funktioner som kan utföras mha Cantor, Gral och Groda kan tillföras på detta sätt.

1. ***Kortaste färdväg***
   Systemet beräknar sträckning och längd för kortaste färdvägen mellan två punkter i vattnet, med hänsyn tagen till hinder i form av öar och grunda vatten.

2. ***Kommunicerande rapporter***
För varje rapport kan avgöras vilka andra rapporter som kan härröra från samma mål (med avseende på tid, lokalisation, ubåtstyp etc).

3. ***Antal ubåtar***
Systemet beräknar troligheten för olika antal ubåtar givet en viss rapportmängd. Man kan också få fram det minsta antal ubåtar som kan ha givit upphov till en viss rapportmängd.

4. ***Evidensområden***
Systemet beräknar med hur stor trolighet det inom ett rektangulärt område, som användaren markerar på kartan, under ett av användaren givet tidsintervall förekommit minst en ubåt. Detta värde baserar sig endast på rapporterna och kan t ex användas för att avgöra när en incident startar, dvs när rapportflödet ej längre kan förklaras av att samtliga rapporter är falska.

5. ***Evidenskartor***
Givet en rapport kan systemet beräkna hur långt den misstänkta ubåten kan ha hunnit under ett visst tidsintervall. På kartan ritas ett färgat utbredningsområde där ubåten kan uppehålla sig. Ju längre tid som förflutit sedan observationen gjordes, ju större blir detta område och ju svagare färgas området på kartan.

   Utbredningsområdena från olika rapporter vägs ihop på basis av evidensteori till en sammanjämkad bild. Resultatet blir en *evidenskarta*, där olika färgnyanser antyder olika grad av trolighet för att en ubåt kan uppehålla sig i området.

   Då en rapport blivit för gammal kommer utbredningsområdet från denna rapport att bli så stort att det inte ger någon information och den tas därför bort ur analysen.

   Denna funktion löper kontinuerligt i tiden och man kan när man så önskar få en bild av nuläget. Man kan även välja att betrakta hur bilden ser ut vid en viss tidpunkt. Det är således även möjligt att presentera evidenskartor över framtida uppehållsområden.

6. ***Troligaste färdväg***
Systemet beräknar troligaste färdvägar fram till nuläget under given hypotes om antal ubåtar. Man kan här tex välja att få de n troligaste färdvägarna presenterade eller samtliga sorterade i trolighetsordning.

   Varje färdväg presenteras tillsammans med ett intervall där den undre intervallgränsen anger till vilken grad rapporterna stöder denna färdväg, medan den övre anger till vilken grad systemet anser att denna färdväg är möjlig (dvs ej motsägs av några kända faktorer).

   De presenterade färdvägarna är ej sanna färdvägar, utan de visar bara längs vilka rapporter de verkliga färdvägarna går. Hur ubåten färdats mellan två rapporter kan systemet f n inte uttala sig om.

   I beräkningarna tar systemet hänsyn till de ingående rapporternas trovärdighet, geografiska läge, tidsavstånd, ubåtstyp och kurser om sådana finns angivna, ubåtars hastigheter, djupdata samt avsaknad av sensorsignaler.

## EVIDENSTEORI

I beräkningarna för de olika analysfunktionerna ingår hantering av osäkra data. Dessa förekommer som osäkerhet hos rapporter (både i form av allmän trovärdighet och i form av osäkerhet i angivelsen av tid och plats), i sensorers detektionsförmåga, i framkomlighet betingad av osäker kunskap om ubåtens djupgående mm.

Dessutom skall osäkerhetsvärden från olika källor vägas samman.

Som tidigare nämnts har vi valt att basera våra osäkerhetsresonemang på evidensteori, som är väl lämpad för den här typen av problem [Lowrance & Garvey, 1983]. Med denna teori som grund är det ej möjligt att beräkna konfidensintervall eller att göra signifikanstest enligt den klassiska teorin. Evidensteorins styrka ligger i dess förmåga att representera och kombinera osäkra data. Vi har därigenom ej tvingats göra stränga och kanske orealistiska antaganden. Det väsentliga har för oss varit att beskriva problemets osäkra data så överskådligt och verklighetstroget som möjligt, för att kunna ge ett så väl beskrivande underlag som möjligt till en beslutsfattare.

Evidensteori baserar sig på att man har *evidenser* (vittnesmål) för en viss hypotes (*positiva evidenser*) samt evidenser mot hypotesen ifråga (*negativa evidenser*). Evidensteorin innehåller regler för hur dessa evidenser skall vägas samman (ortogonal kombination, *Dempster's regel*) [Shafer, 1976]. En skillnad mot den klassiska sannolikhetsteorin är att om vi för en hypotes har en evidens som tilldelats värdet p, betyder detta ej att man misstror hypotesen till värdet 1-p. Om det exempelvis inkommit en rapport om en ubåt som givits tilltron p = 0.6, betyder detta ej att vi misstror att det kan ha förekommit en ubåt i området till graden 0.4. Vi saknar helt enkelt kunskap att fördela hela massan 1 och förbehåller oss rätten att vara ovetande till graden 0.4. Vi har ju i detta fall inte något konkret indicium som talar mot en ubåt. Man skiljer alltså i evidensteorin på det som talar mot en hypotes och på det man saknar kunskap om.

Resultatet fås i form av ett intervall där det lägre värdet anger det stöd evidenserna ger hypotesen och det högre värdet anger till vilken grad evidenserna ej motsäger hypotesen, dvs till vilken grad hypotesen kan anses möjlig.

En mycket viktig förutsättning här är att de olika evidenskällorna kan betraktas som oberoende av varandra. Vid sammanvägningen av evidenser multipliceras de olika trolighetsvärdena ihop, vilket kan leda till helt felaktiga slutsatser om oberoendeantagandet ej är uppfyllt. Exempelvis kan två signaler från samma sensor inom ett kort tidsintervall ej betraktas som oberoende.

I vår tillämpning förekommer positiva evidenser i form av rapporter samt negativa evidenser i form av uteblivna sensorsignaler och dålig framkomlighet (t ex för grunt vatten).

Vid beräkning av troligaste färdväg kan tiden mellan två rapporter vara i knappaste laget, även om det inte är en omöjlighet att ta sig fram på denna tid. Tidsfaktorn blir då en negativ evidens med ett p-värde som är proportionellt mot tidsfaktorn.

## ÖVERSIKT ÖVER ALGORITMER

***Definition.*** Låt [p, P], $0 \leq p \leq P \leq 1$, vara ett evidensintervall för en hypotes **A**. Då är p stödet för **A** och P den grad till vilken **A** är möjlig.

En grundläggande analysmetod är att beräkna den *kortaste vägen* mellan två rapporter. Detta sker genom sökning i en *visibility graph* [Asano *et al.*, 1986] konstruerad utifrån öars eller djupkurvors *relativa konvexa höljen* [Toussaint, 1985]. Analysmetoden används både separat och som del i andra metoder.

De flesta analysfunktioner bygger på en värdering av kopplingar mellan par av rapporter. Värderingen av en sådan koppling beror på ett flertal faktorer som var och en, till en viss grad, talar emot en koppling, dvs de har ett evidensintervall på formen [0, P].

Den viktigaste faktorn är hastighetskravet för att förflytta sig enligt den *kortaste vägen* mellan två rapporter, i jämförelse med främmande ubåtars antagna hastigheter. Efter det att den kortaste vägen har beräknats kan hastighetskravet beräknas utifrån vägsträckningen och de ingående rapporternas tider. Övriga faktorer som påverkar värderingen av en koppling är de sensorer, med sina upptäcktssannolikheter, vilka ligger mellan de två rapporterna men ej har lämnat någon signal, samt eventuella kursdifferenser och iakttagelser om ubåtstyp. Alla dessa faktorer utvärderas efter speciella regler med hänsyn tagen till respektive faktors betydelse. Slutligen kan en sammanvägning av de olika faktorerna ske med en ortogonal kombination.

Analysmetoden *kommunicerande rapporter* bygger på att värden för alla kopplingarna beräknas enligt ovan. Därefter plottas de kopplingar till en vald rapport vars värde är större än ett visst gränsvärde.

Analysmetoden *antal ubåtar* ger ett evidensintervall för alla alternativa antal, från ingen ubåt till det minsta antal ubåtar som krävs om alla rapporter är sanna. Metoden bygger på en undersökning av den ovan beskrivna grafen av kopplingar, $\{q_{ij}\}$. De inbördes separerade rapporterna (koppling saknas) ger trolighetsintervall för alla alternativ beträffande antalet ubåtar.

*Evidensområden* baseras enbart på rapporter. Rapporter har ett evidensintervall på formen [p, 1], där värdet p sätts genom en tolkning av rapporten. Rapporterna, $r_i$, betraktas som en evidens för främmande undervattensverksamhet utan åtskillnad mellan olika ubåtar. Genom en ortogonal kombination beräknas ett evidensintervall för att det finns åtminstone en ubåt inom den definierade rektangeln, Resultat blir $[1 - \sum_{\forall i | r_i \in R} (1 - p_i)\ , 1]$.

En analysmetod skiljer sig från de ovannämnda, nämligen *evidenskartor*. De ovannämnda metoderna behandlar förflyttningar mellan rapporter medan evidenskartor presenterar lägesbilder med trolighet mellan rapporters tider. Metoden utgår ifrån de inkommande rapporterna och fasta bakgrundsdata. Havet är uppdelat i rutor, 500x500 meter, och varje inkommande rapports trolighet placeras i den ruta där rapporten är placerad. Allt eftersom tiden går flyttas en del av troligheten till angränsande rutor, dock med hänsyn tagen till djupkurvor. Om trolighet flyttas till en ruta vari en sensor kan upptäcka en ubåt men inte givit någon signal, sker en ortogonal kombination mellan troligheten och sensorns upptäcktssannolikhet, varefter trolighet för ubåt i rutan samt den trolighet som kommer att flyttas till andra sidan sensorn sjunker högst markant. Vid varje tidsintervall sker en ortogonal kombination i varje ruta av de troligheter som härrör sig från olika rapporter. Resultatet blir en evidenskarta som visar, för alla rutor, sannolikheten att finna minst en ubåt i rutan.

En analysmetod som beror både på rapporterna och kopplingarna är *troligaste färdväg*. Denna metod bygger på att först sätta upp de tänkbara färdvägarna varefter troligheten för varje färdväg beräknas. Vi kan hitta alla tänkbara färdvägar genom att betrakta grafen av alla kopplingar där $P > 0$.

Exempel:

Låt oss lösa problemet för fallet två rapporter, $r_1$ och $r_2$, och <u>en</u> möjlig ubåt:

***Definition.*** Låt $\langle \ldots, r_i, \ldots \rangle$, $\langle \ldots, \neg r_i, \ldots \rangle$ och $\langle \ldots, \theta_i, \ldots \rangle$ vara representationerna för ubåt vid $r_i$, ingen ubåt vid $r_i$ resp. ingen kunskap om $r_i$, samt låt $\{\langle \ldots, r_i, \ldots \rangle, \langle \ldots, r_j, \ldots \rangle\}$ betyda $\langle \ldots, r_i, \ldots \rangle$ eller $\langle \ldots, r_j, \ldots \rangle$.

Då representerar $\langle r_1, \theta \rangle$, $\langle \theta, r_2 \rangle$ och $\{\langle \neg r_1, \theta \rangle, \langle \theta, \neg r_2 \rangle\}$ ubåt vid $r_1$ resp. $r_2$ samt inte både $r_1$ och $r_2$ dvs ingen koppling $r_1 \rightarrow r_2$.
Vi har då fyra tänkbara *färdvägar*:

$\langle r_1, r_2 \rangle$ $\equiv$ ubåten har varit vid $r_1$ och åkt till $r_2$.
$\langle r_1, \neg r_2 \rangle$ $\equiv$ ubåten har varit vid $r_1$, $r_2$ är falsk.
$\langle \neg r_1, r_2 \rangle$ $\equiv$ ubåten har varit vid $r_2$, $r_1$ är falsk.
$\langle \neg r_1, \neg r_2 \rangle$ $\equiv$ båda rapporterna är falska.

Låt $p_1$ och $p_2$ vara stöden för rapporterna och $q_{12}$ vara stödet för <u>ingen</u> koppling mellan $r_1$ och $r_2$ (dvs vi har evidensintervallen $[p_1, 1]$ och $[p_2, 1]$ för rapporterna samt $[0, 1 - q_{12}]$ för kopplingen $r_1 \rightarrow r_2$).

Vi kan nu utföra de ortogonala kombinationerna (*Dempsters regel*) mellan de tre evidenserna, $r_1$, $r_2$ och $r_1 \rightarrow r_2$. Först utför vi kombinationen mellan $r_1$ och $r_2$, $r_1 \oplus r_2$:

| | $\langle \theta, r_2 \rangle$ <br> $p_2$ | $\langle \theta, \theta \rangle$ <br> $1 - p_2$ |
|---|---|---|
| $\langle r_1, q \rangle$ <br> $p_1$ | $\langle r_1, r_2 \rangle$ <br> $p_1 \cdot p_2$ | $\langle r_1, \theta \rangle$ <br> $p_1 \cdot (1 - p_2)$ |
| $\langle \theta, \theta \rangle$ <br> $1 - p_1$ | $\langle \theta, r2 \rangle$ <br> $(1 - p_1) \cdot p_2$ | $\langle \theta, \theta \rangle$ <br> $(1 - p_1) \cdot (1 - p_2)$ |

Därefter utför vi kombinationen av ovanstående resultat, $r_1 \oplus r_2$, och $r_1 \rightarrow r_2$, $r_1 \oplus r_2 \oplus r_1 \rightarrow r_2$:

| | $\{<\neg r_1, \theta>, <\theta, \neg r_2>\}$<br>$q_{12}$ | $<\theta, \theta>$<br>$1 - q_{12}$ |
|---|---|---|
| $<r_1, r_2>$<br>$p_1 \cdot p_2$ | $\Phi$<br>$p_1 \cdot p_2 \cdot q_{12}$ | $<r_1, r_2>$<br>$p_1 \cdot p_2 \cdot (1 - q_{12})$ |
| $<r_1, \theta>$<br>$p_1 \cdot (1 - p_2)$ | $<r_1, \neg r_2>$<br>$p_1 \cdot (1 - p_2) \cdot q_{12}$ | $<r_1, \theta>$<br>$p_1 \cdot (1 - p_2) \cdot (1 - q_{12})$ |
| $<\theta, r_2>$<br>$(1 - p_1) \cdot p_2$ | $<\neg r_1, r_2>$<br>$(1 - p_1) \cdot p_2 \cdot q_{12}$ | $<\theta, r_2>$<br>$(1 - p_1) \cdot p_2 \cdot (1 - q_{12})$ |
| $<\theta, \theta>$<br>$(1 - p_1) \cdot (1 - p_2)$ | $\{<\neg r_1, \theta>, <\theta, \neg r_2>\}$<br>$(1 - p_1) \cdot (1 - p_2) \cdot q_{12}$ | $<\theta, \theta>$<br>$(1 - p_1) \cdot (1 - p_2) \cdot (1 - q_{12})$ |

Det första vi kan konstatera är att vi har erhållit en konflikt, $\Phi$. Den faktorn, $p_1 \cdot p_2 \cdot q_{12}$, måste vi normera bort i de fortsatta beräkningarna. För att erhålla evidensintervallen ska vi summera alla de bidrag som stöder resp. möjliggör de olika färdvägarna. Färdvägen $<r_1, r_2>$ stöds av bidraget från rad 1 kolumn 2, r1k2, samt möjliggörs av r1k2, r2k2, r3k2 och r4k2.

Dess evidensintervall blir: $\left[\frac{p_1 \cdot p_2 \cdot (1 - q_{12})}{1 - p_1 \cdot p_2 \cdot q_{12}}, \frac{1 - q_{12}}{1 - p_1 \cdot p_2 \cdot q_{12}}\right]$

På samma sätt kan övriga färdvägars evidensintervall beräknas:

$<r_1, \neg r_2>$: r2k1 resp. r2k1, r4k1, r2k2 och r4k2 $\Rightarrow$ $\left[\frac{p_1 \cdot (1 - p_2) \cdot q_{12}}{1 - p_1 \cdot p_2 \cdot q_{12}}, \frac{1 - p_2}{1 - p_1 \cdot p_2 \cdot q_{12}}\right]$ ,

$<\neg r_1, r_2>$: r3k1 resp. r3k1, r4k1, r3k2 och r4k2 $\Rightarrow$ $\left[\frac{(1 - p_1) \cdot p_2 \cdot q_{12}}{1 - p_1 \cdot p_2 \cdot q_{12}}, \frac{1 - p_1}{1 - p_1 \cdot p_2 \cdot q_{12}}\right]$ ,

$<\neg r_1, \neg r_2>$: inga resp. r4k1 och r4k2 $\Rightarrow$ $\left[0, \frac{(1 - p_1) \cdot (1 - p_2)}{1 - p_1 \cdot p_2 \cdot q_{12}}\right]$

Ett *konfliktmått*, k, erhålls vid beräkning av troligaste färdväg. I beräkningen ovan, $r_1 \oplus r_2 \oplus r_1 \rightarrow r_2$, uppstod en konflikt i ruta r1k1. Konfliktmåttet, $0\% \leq k \leq 100\%$, är i detta fall bidraget från r1k1, dvs $100 \cdot p_1 \cdot p_2 \cdot q_{12}\%$.

## MÖJLIGA FRAMTIDA UTVECKLINGAR

### Bättre interaktivitet

Dezzy uppvisar f n vissa brister när det gäller användarens möjligheter att enkelt styra sitt scenario. Användaren kan t ex själv vilja konstruera en framkomlighetskarta, genom att klassificera godtyckliga områden på kartan i olika grupper.

Grupperna graderas efter hur troligt användaren anser det vara att en ubåt med hänsyn till framkomligheten kunnat ta sig fram genom sådana områden (eller valt att ta sig fram). Systemet ritar då upp en karta med olikfärgade områden för olika framkomlighetsgrad.

Hur ett område klassificeras kan bero på kännedom om taktiska regler, egen uppfattning om navigeringssvårighet o d. Det kan också vara fallet att man noga genomsökt ett område med rörliga sensorer så att detta område för en tid skall betraktas som mindre troligt.

En annan intressant generalisering vore att möjliggöra ansättande av olika maximal hastighet i olika punkter eller områden. Snabbaste väg blir då inte längre ekvivalent med kortaste väg.

### Metoder för känslighetsanalys och sammanfattning

En angelägen vidareutveckling av Dezzy-AM är införande av metoder för känslighetsanalys och sammanfattning (*gisting*) [Lowrance & Strat, 1988]. Med hjälp av metoder för känslighetsanalys kan en användare snabbt ta reda på vilka indata-evidenser som bidrar väsentligt till ett visst resultatintervall. Metoder för sammanfattning går ut på att systemet halvautomatiskt slår ihop en mängd möjliga alternativa händelser till en "*superhändelse*", som i fortsättningen betraktas som odelbar, dvs man intresserar sig inte längre för troligheten för dess komponenthändelser. Någon sådan teknik är i praktiken nödvändig för att man skall kunna hantera den stora mängd alternativ som mycket ofta uppkommer vid användning av evidenskalkyl. I Dezzy uppstår sådana problem tex vid beräkning av troligaste färdväg (men inte vid beräkning av evidenskartor).

### Metoder för prediktion

En möjlighet att prediktera ubåtars framtida lägen vore ett väsentligt tillskott till Dezzy. Detta kan uppnås med åtminstone två olika principer:

(*i*) ***Mönsterigenkänning***
Genom att analysera historiska data kan förekommande rapportsekvenser delas in i klasser. Klassificeringsprincipen kan fastställas antingen manuellt eller automatiskt. Det finns ett flertal olika matematiska metoder som kan komma ifråga om man vill göra en sådan klassificering. En aktuell situation kan sedan automatiskt hänföras till en viss klass. Jämförelse med andra rapportsekvenser som tillhör samma klass kan nu ge användbar information.

(*ii*) ***Taktisk kunskap***
Genom intervjuer med militära ämnesexperter kan en taktisk kunskapsbas byggas upp. En sådan kunskapsbas skulle innehålla information om ubåtars beteende under olika skeenden av en ubåtsoperation samt information om ubåtars tänkbara målsättningar. Informationen i denna kunskapsbas är avsevärt mindre detaljerad än informationen i de historiska mönstren. Å andra sidan kan kunskap på en "högre" nivå lagras.

Två av de tidigare beskrivna analysmetoderna skulle vinna på att använda prediktion:

(*i*) ***Troligaste färdväg***
Analysmetoden troligaste färdväg skulle kunna kompletteras med prediktion. De olika färdvägarna skulle då kunna ge information om framtida troliga lägen samt kurser efter de sista rapporterna, medan de idag endast behandlar situationen fram till den sista rapporten.

(*ii*) ***Evidenskartor***
Prediktionen skulle kunna vägas in i evidenskartor som har beräknats för en framtida tidpunkt. Till skillnad från fallet troligaste färdväg skulle detta inte ge detaljkunskap om de olika tänkbara fallen utan i stället en mer sammanfattande bild av ett framtida läge.

### Metoder för dimensionering och värdering av underrättelsesystem

Genom att utveckla det beskrivna informationssystemet bör man kunna få hjälp vid konfigurering och värdering av det egna underrättelsesystemet. Idén är att använda analyssystemet som en modul i en simulator, som genom att köra många spel efter någon systematisk princip, ger statistik över troligheten för upptäckt, lokalisering och eventuellt följning.

För en given uppsättning sensorer (fasta och rörliga), en given miljö och en given rörlig "hotbild" (banor för en eller flera farkoster med definierade utstrålningsegenskaper), beräknas täcknings-områden samt upptäckts-, lokaliserings- och följningstrolighet under olika taktiska förutsättningar.

Hypoteser om hotbild, miljö och taktik hos målet skall kunna varieras. Användaren bör enkelt kunna ändra sensorbilden inför ett nytt spel och eventuellt bör systemet kunna optimera utplaceringen av sensorer (antal, lokalisering).

## SLUTSATSER

Arbetet med Dezzy har visat att evidenskalkyl är en användbar metodik vid analys av ubåtsunderrättelser, åtminstone i fredstida scenarier som omfattar många observationer.

Arbetet har lett till att flera praktiskt användbara metoder utvecklats. För vissa problem återstår dock mycket arbete innan evidenskalkylen kan få praktisk tillämpning.

Som vanligt när en ny, formell metodik introduceras inom ett tillämpningsområde måste ett visst teoretiskt kunnande och en praktisk erfarenhet byggas upp bland användarna innan den nya metodiken kan få verklig betydelse.

## ERKÄNNANDE

## REFERENSER

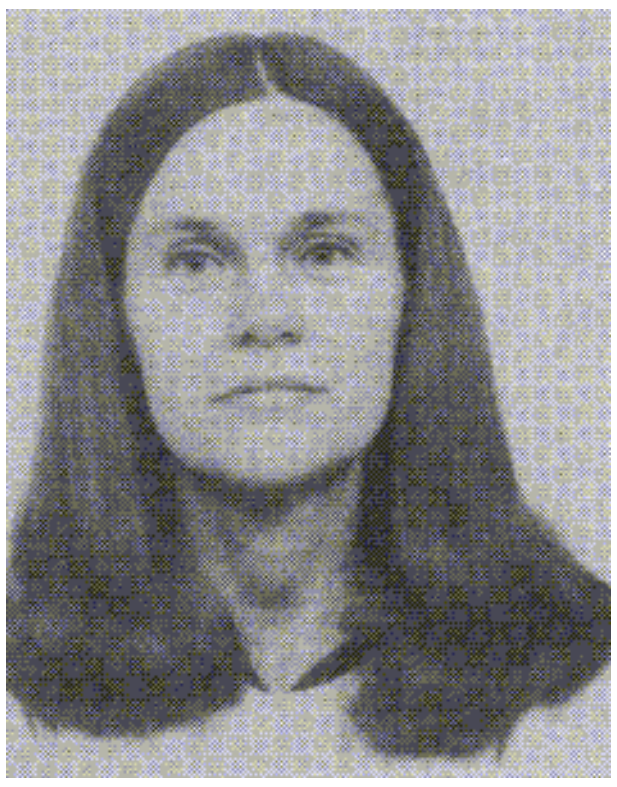

*Ulla Bergsten* är matematisk statistiker och avdelningsdirektör vid Försvarets forskningsanstalt, där hon varit anställd sedan 1983. Hon erhöll fil.kand. examen 1974 och bedrev därefter forskarstudier i matematisk statistik. Ulla Bergsten var under åren 1972 - 1983 anställd vid institutionen för matematisk statistik, Stockholms Universitet. Där arbetade hon som lärare och forskare samt som konsult inom statistiska forskningsgruppen. Hon har vid Försvarets forskningsanstalt arbetat med databasteknik för statistiska tillämpningar samt med metoder för analys av data, främst evidenskalkyl, i beslutsstöds- och underrättelsesystem.

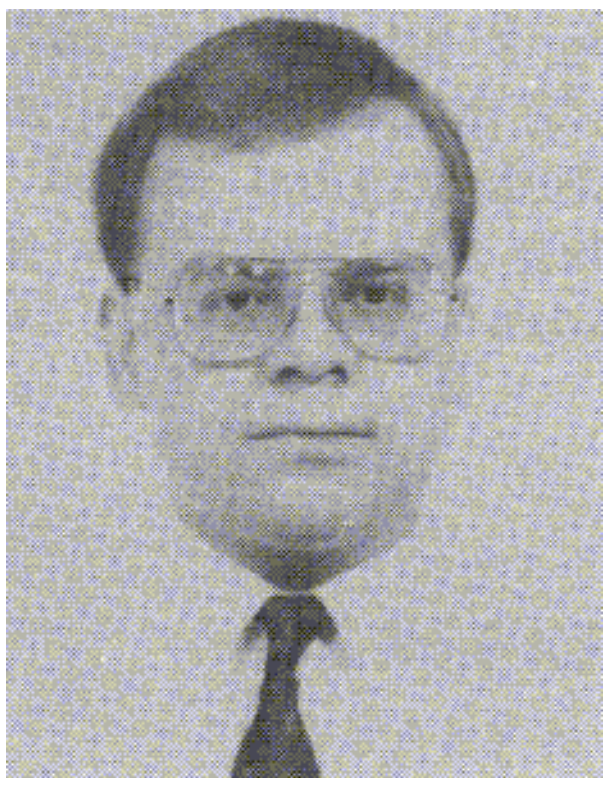

*Johan Schubert* är forskare vid Försvarets forskningsanstalt sedan 1987. Han erhöll sin civilingenjörsexamen i teknisk fysik vid KTH 1986. För närvarande forskar han kring artificiell intelligens och evidenskalkyl i militära beslutsstöds- och underrättelsesystem.

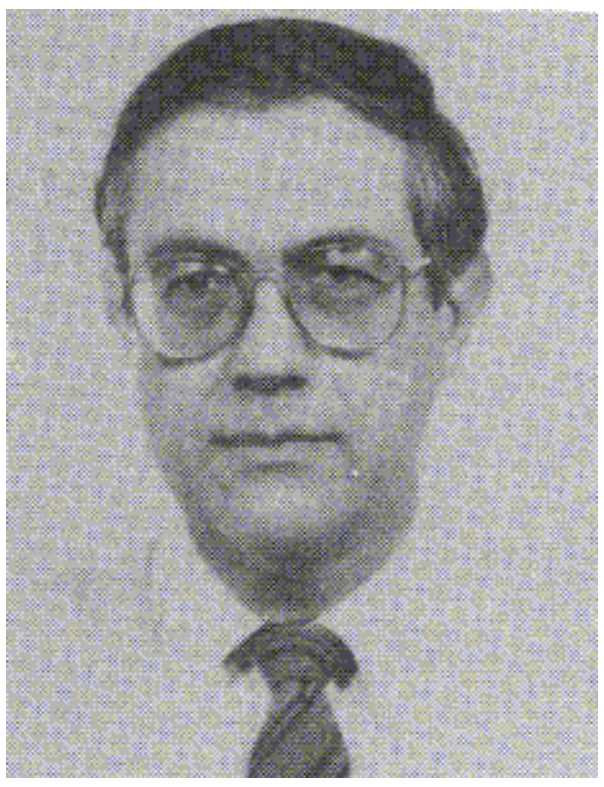

*Per Svensson* är civilingenjör, teknisk fysik, KTH 1965, teknisk licentiat i informationsbehandling, särskilt numerisk analys KTH 1970, samt teknisk doktor i informationsbehandling, särskilt numerisk analys KTH 1979. Han antogs som oavlönad docent i informationsbehandling vid KTH 1984. Han är sedan 1987 forskningschef i datavetenskap vid Försvarets forskningsanstalt. Per Svensson har tidigare varit anställd vid KTH:s institution för informationsbehandling 1964 - 1969 samt vid IBM Svenska AB 1969 - 1973. Hans forskning har främst rört databasteknik för tekniskt - vetenskapliga tillämpningar. På senare tid har han även arbetat med teknik för militära lednings- och underrättelsesystem samt med geografiska data- och kunskapsbaser.

## POST-CONFERENCE APPENDIX. DEZZY PICTURES (Unpublished)

DEZZY is a decision support system for anti-submarine intelligence analysis using Dempster-Shafer theory. The aim of the system is through situation assessment based on geographical knowledge, knowledge of the intelligence forces whereabouts and a quantitative analysis of available intelligence reports offer decision support to intelligence analysts.

Since every intelligence report contains information such as time, position, velocity, direction and submarine type that might be uncertain and in fact every intelligence report itself comes with a general uncertainty classification, the ability to reason under uncertainty is of the utmost importance.

A number of analysis methods based on uncertainty reasoning are available to the intelligence analyst. The two most important are methods for locating and tracking submarines. Apart from the evidence in the intelligence reports these methods also take into account, for example, the "negative evidence" in geographical distance, deep-draughts and non-firing sensors.

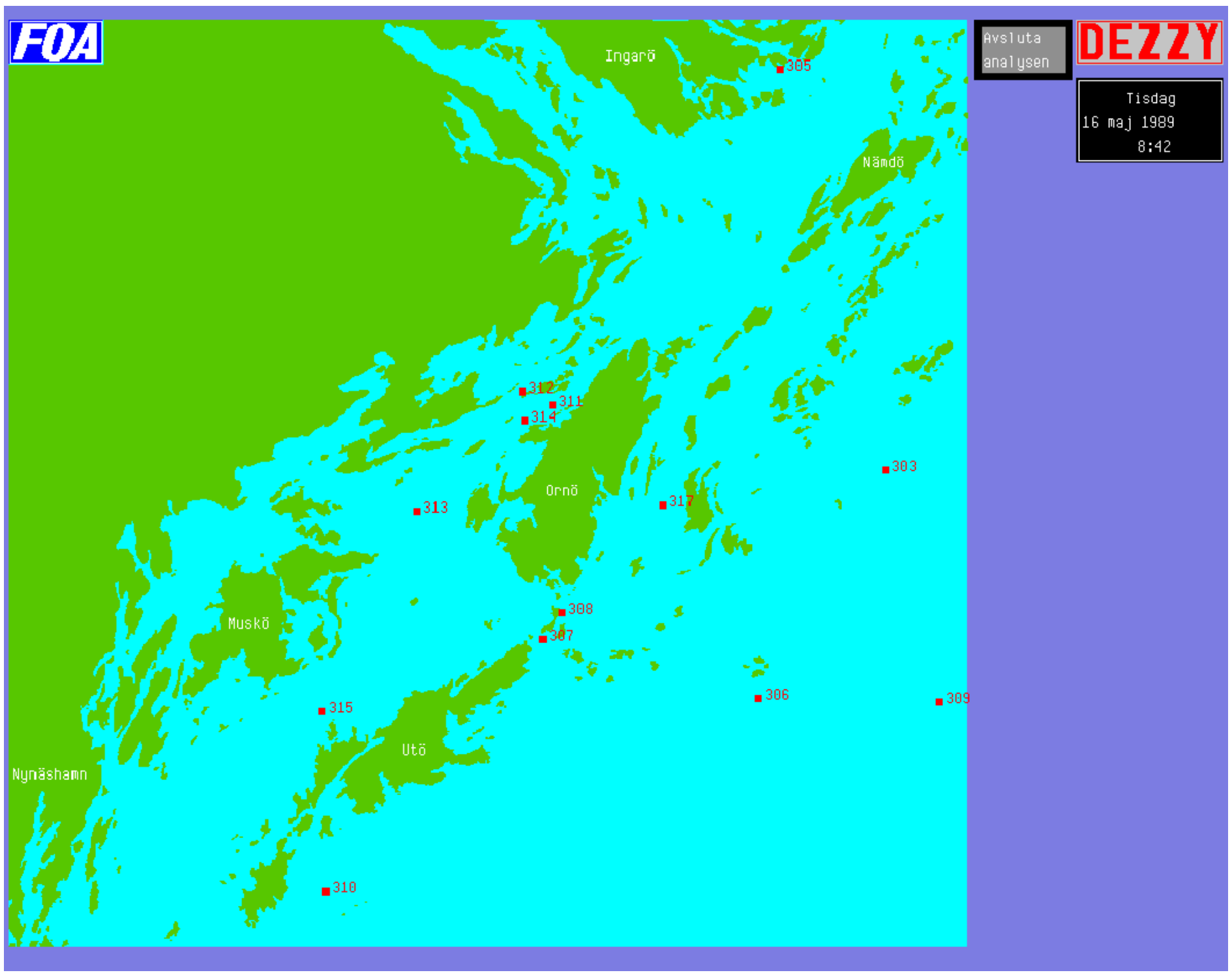

**Figure 1.** A scenario of intelligence reports (red) in the south archipelago of Stockholm (simulated data).

### Locating submarines

In support of locating submarines an evidence map is created where different areas are classified according to their plausibility of containing a submarine. This is done by fusing all intelligence reports with their respectively possible whereabouts at a chosen time. For every subarea the plausibility is then calculated from the result of the fusion in order to create the evidence map.

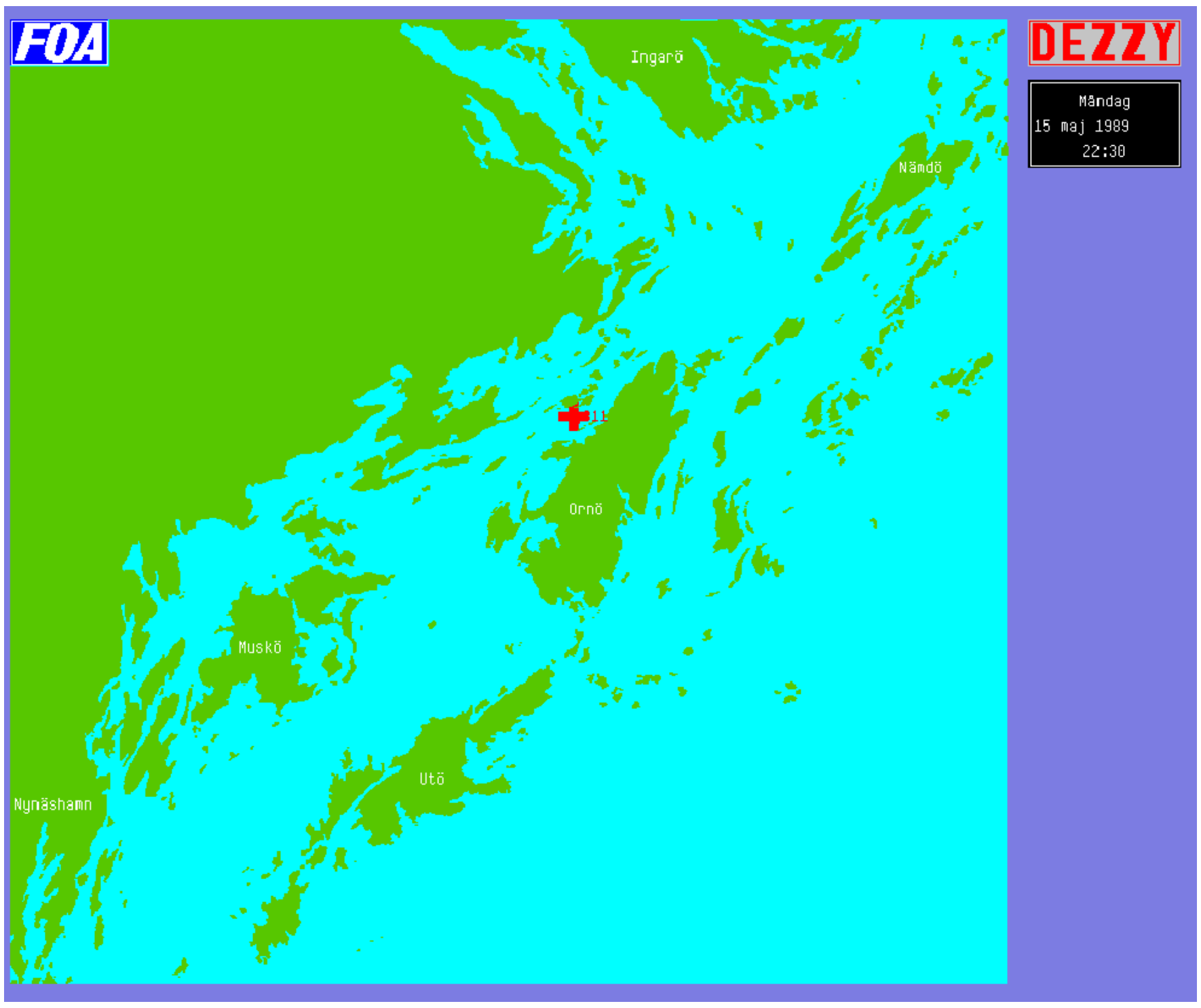

**Figure 2.** The initial contact at 2230 hours.

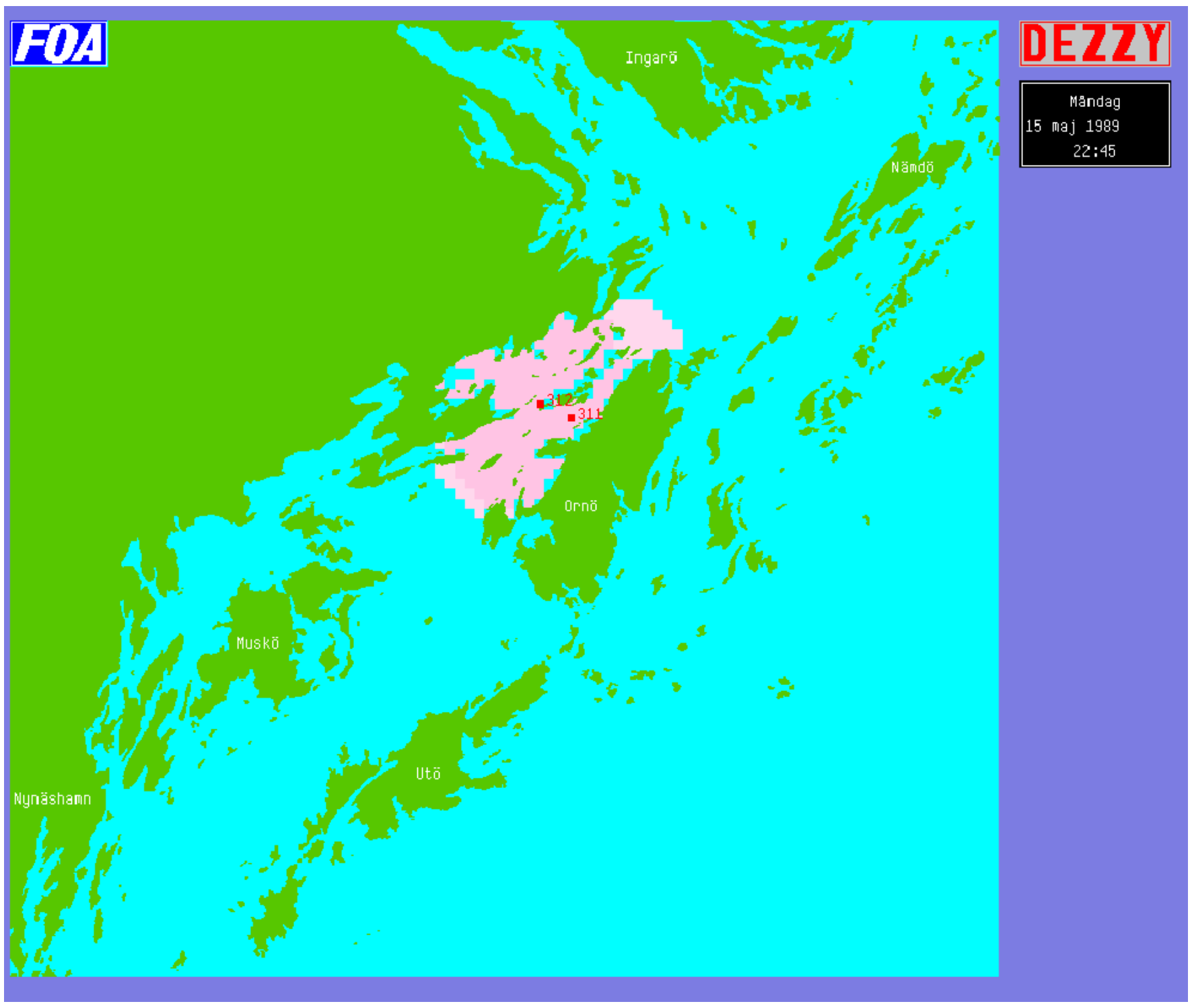

**Figure 3.** 15 minutes later.

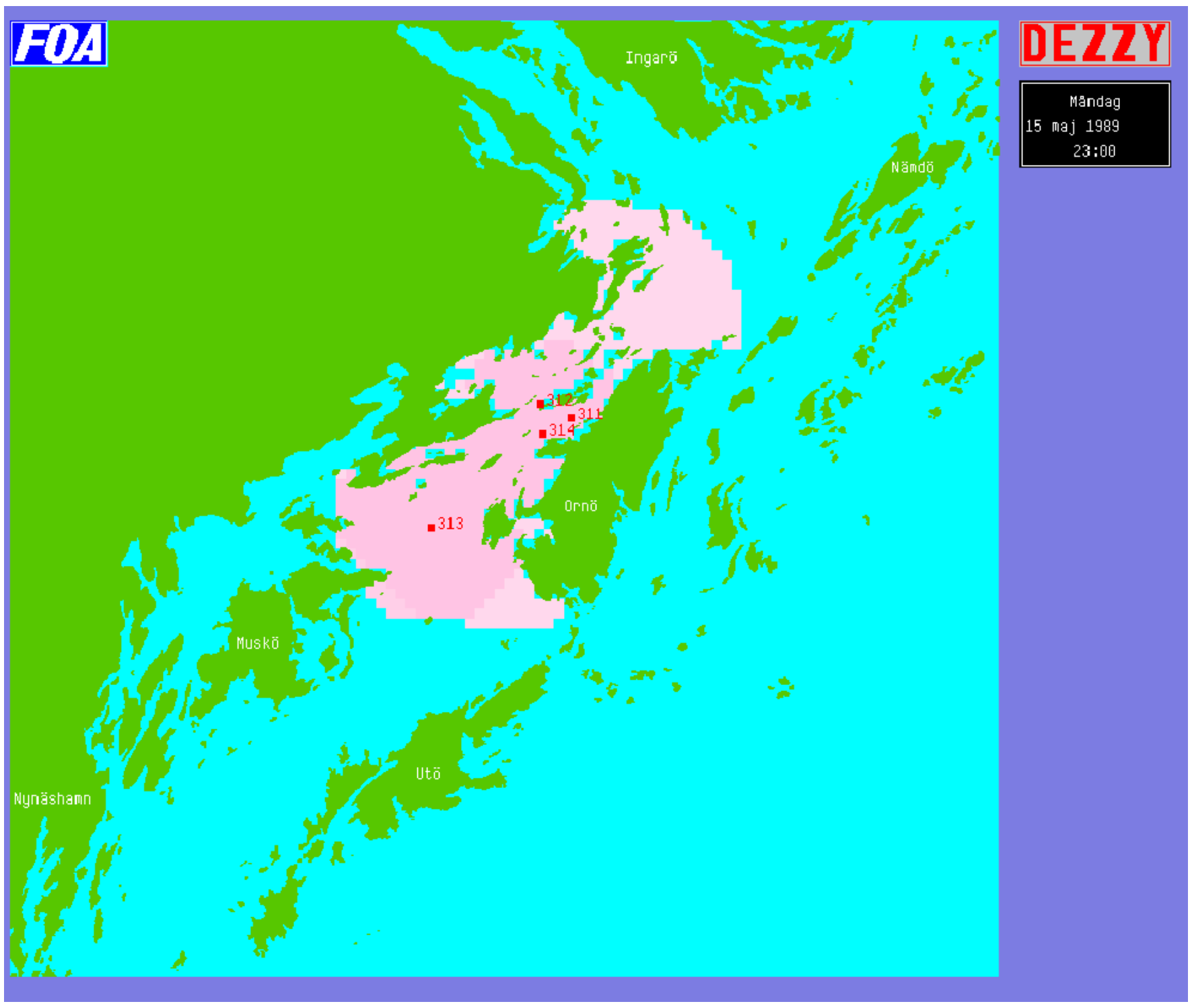

**Figure 4.** At 2300 hours.

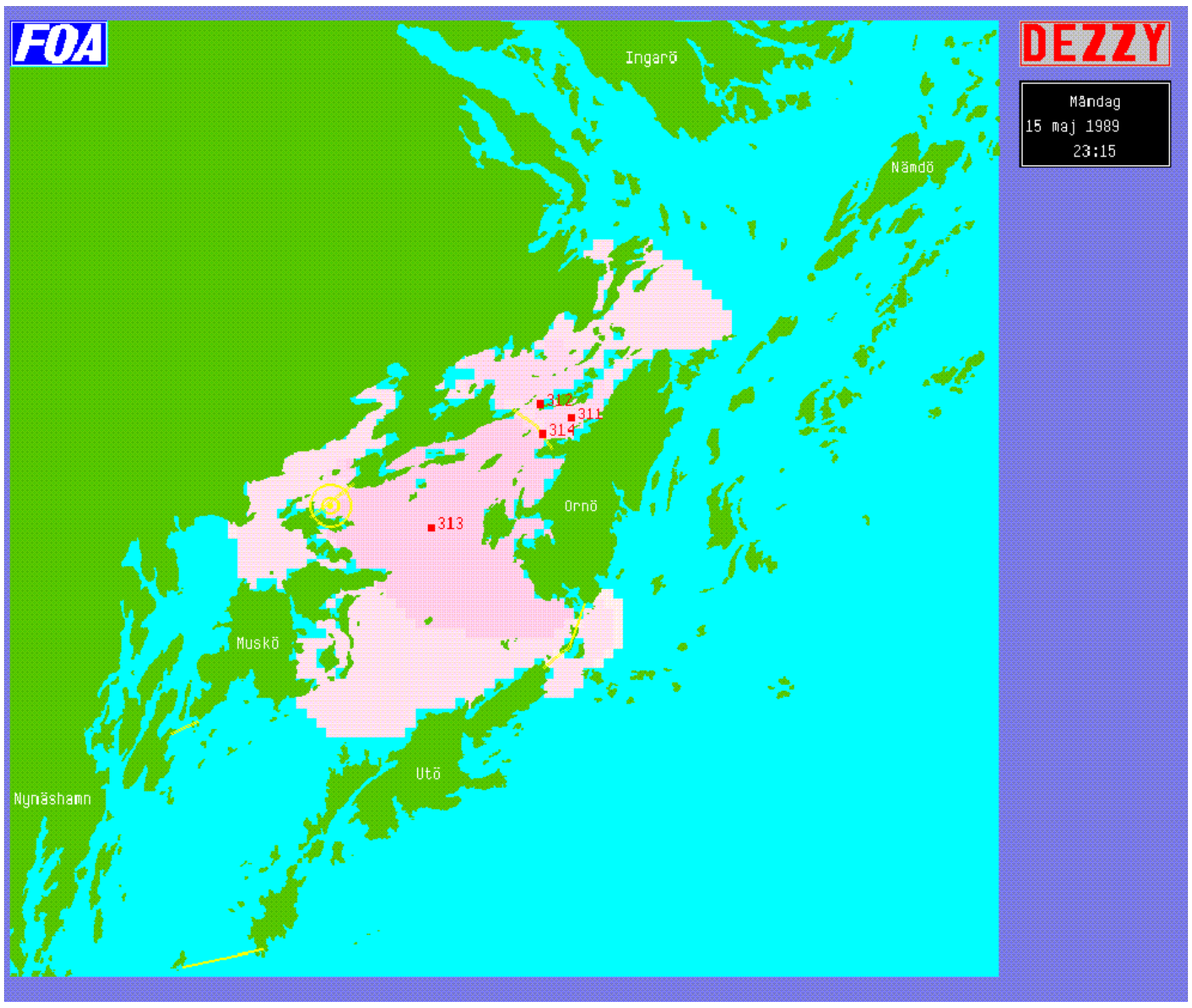

**Figure 5.** Final evidence map at 2315 hours.

### Tracking submarines

In support of the tracking of submarines a ranking is made according to the probabilities of all possible paths through a directed acyclic graph of intelligence reports. When calculating the belief and plausibility of a path account is taken both to the positive evidence of the intelligence report and to the negative evidence against transitions between reports. The intelligence analyst is able to interact with the system through a series of menus and geographical maps and is able to test different hypothesis by selection and manipulation of intelligence reports and sensors.

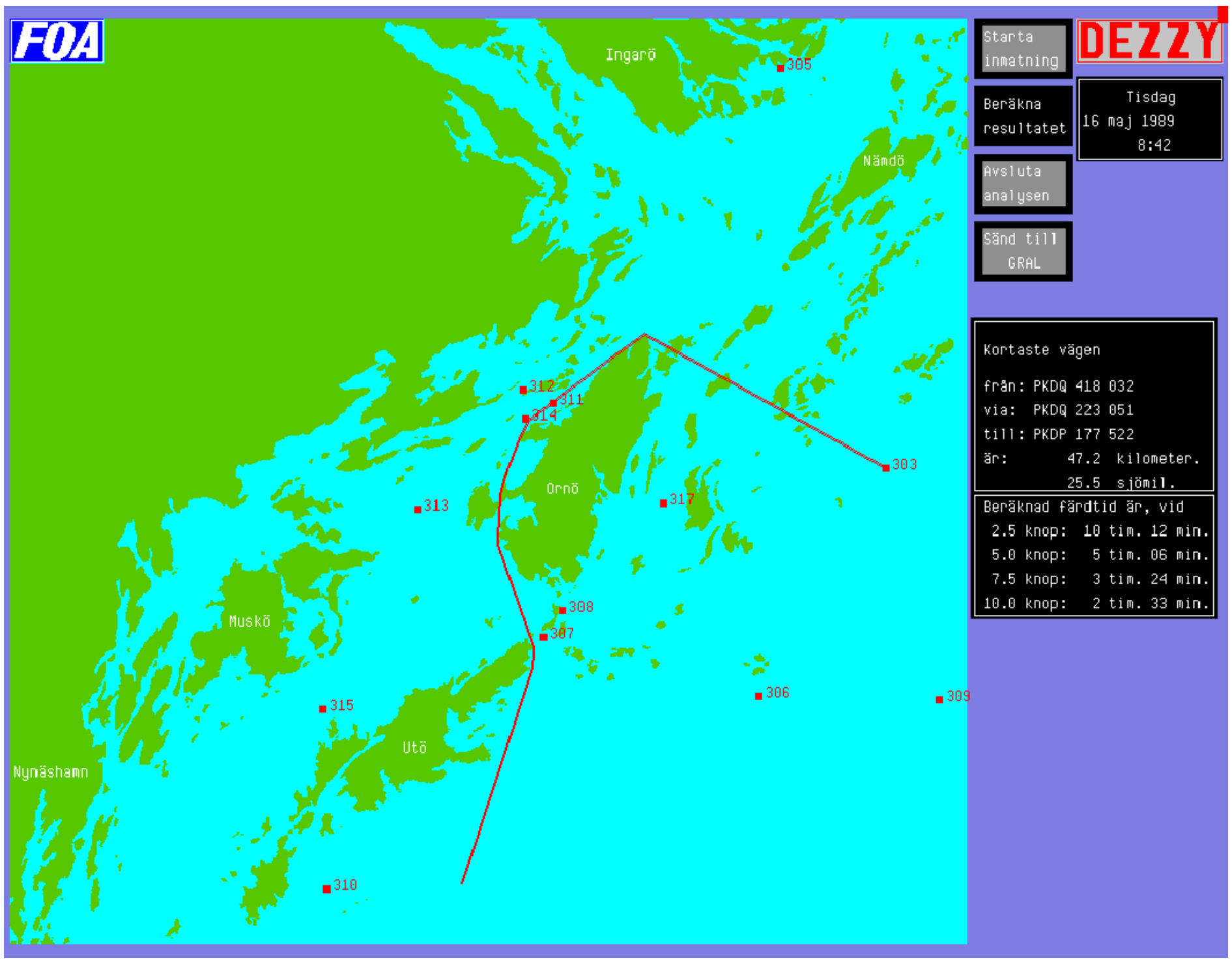

**Figure 6.** Intelligence reports are connected by shortest path using the A*-algorithm.

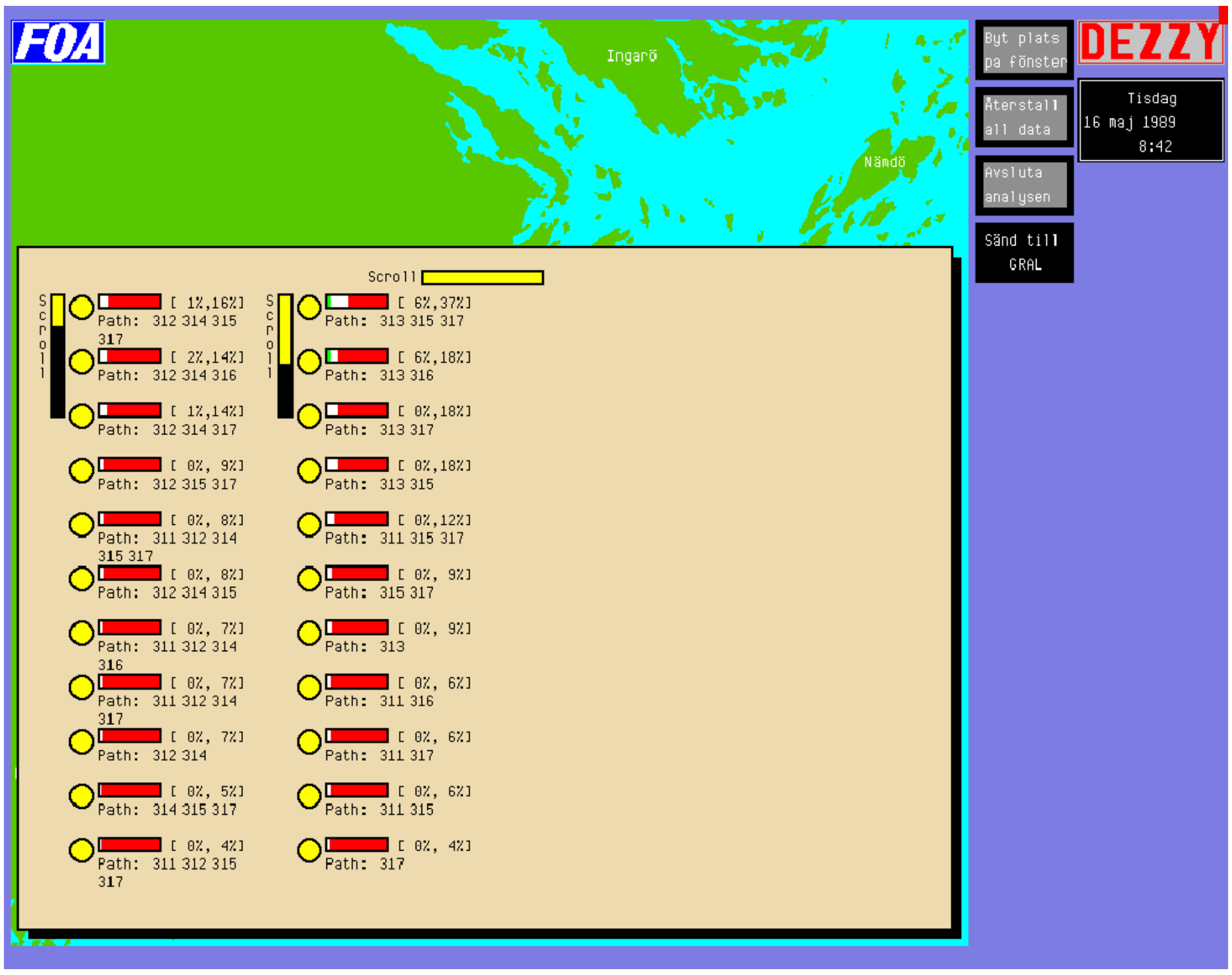

**Figure 7.** Two submarines are identified (each column) with many different possible paths (each row).

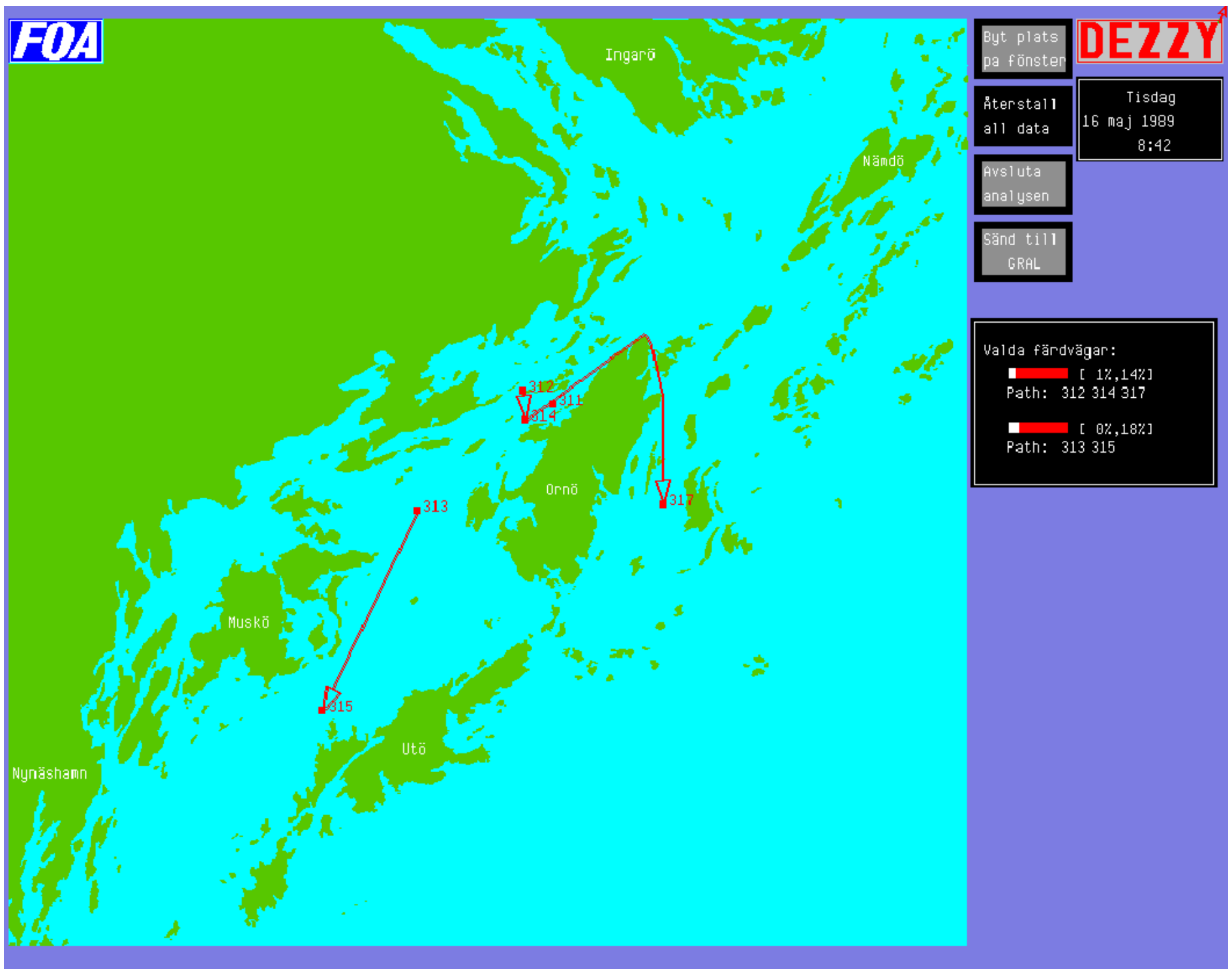

**Figure 8.** The analyst investigates different possibilities before making his choice. Different alternatives are plotted in the map.